\documentclass[a4paper, 10pt, conference]{ieeeconf}      
\pdfoutput=1
\usepackage[utf8]{inputenc} 

\IEEEoverridecommandlockouts                              

\usepackage{xurl}
\usepackage{graphicx}
\usepackage{stfloats}
\usepackage{amsmath}
\usepackage{amsfonts}
\usepackage{algorithm}
\usepackage{algpseudocode}
\usepackage{booktabs}
\usepackage{hhline}
\usepackage[table]{xcolor}
\usepackage{multirow}
\usepackage{array}
\usepackage{balance}

\makeatletter
\let\NAT@parse\undefined
\makeatother

\usepackage[colorlinks=true, linkcolor=blue, citecolor=blue, urlcolor=blue, hypertexnames=false]{hyperref}
\usepackage[all]{hypcap}

\usepackage{cite}

\usepackage[nameinlink,noabbrev]{cleveref}
\crefname{figure}{Fig.}{Figs.}
\crefname{equation}{Eq.}{Eqs.}
\crefname{section}{Sec.}{Secs.}
\crefname{table}{Table}{Tables}

\definecolor{lightred}{RGB}{255,150,150}
\definecolor{lightgreen}{RGB}{144,238,144}
\definecolor{lightgray}{RGB}{211,211,211}

\title{\LARGE \bf
SAATT Nav: a Socially Aware Autonomous Transparent Transportation Navigation Framework for Wheelchairs}

\author{Yutong Zhang$^{1*}$, Shaiv Y. Mehra$^{2*}$, Bradley S. Duerstock$^{1, 2}$, Juan P. Wachs$^{1, 2}$
\thanks{*Equal contribution.}
\thanks{$^{1}$Edwardson School of Industrial Engineering, Purdue University, West Lafayette, IN 47907, USA (e-mail: \{zhan3606, bsd, jpwachs\}@purdue.edu).}
\thanks{$^{2}$Weldon School of Biomedical Engineering, Purdue University, West Lafayette, IN 47907, USA (e-mail: mehra19@purdue.edu).}
\thanks{Corresponding author: Yutong Zhang (zhan3606@purdue.edu).}
}

\begin{document}

\maketitle
\thispagestyle{empty}
\pagestyle{empty}

\begin{abstract}
While powered wheelchairs reduce physical fatigue as opposed to manual wheelchairs for individuals with mobility impairment, they demand high cognitive workload due to information processing, decision making and motor coordination. Current autonomous systems lack social awareness in navigation and transparency in decision-making, leading to decreased perceived safety and trust from the user and others in context. This work proposes Socially Aware Autonomous Transparent Transportation (SAATT) Navigation framework for wheelchairs as a potential solution. By implementing a Large Language Model (LLM) informed of user intent and capable of predicting other peoples’ intent as a decision-maker for its local controller, it is able to detect and navigate social situations, such as passing pedestrians or a pair conversing. Furthermore, the LLM textually communicates its reasoning at each waypoint for transparency. In this experiment, it is compared against a standard global planner, a representative competing social navigation model, and an Ablation study in three simulated environments varied by social levels in eight metrics categorized under Safety, Social Compliance, Efficiency, and Comfort. Overall, SAATT Nav outperforms in most social situations and equivalently or only slightly worse in the remaining metrics, demonstrating the potential of a socially aware and transparent autonomous navigation system to assist wheelchair users.
\end{abstract}


\section{INTRODUCTION}

Approximately 80 million people worldwide require a wheelchair for mobility, spanning older adults and individuals with neurodegenerative diseases, spinal cord injuries, stroke, and other conditions. This population is growing as medical advancements improve survival rates across age groups~\cite{who2023}. While powered wheelchairs reduce physical strain endured by those using manual wheelchair~\cite{kang2025}, users still face significant cognitive load from environmental perception, navigation, and control. Neuroimaging studies using fNIRSs show that navigating cluttered environments increases prefrontal cortex activation, reflecting higher attentional and executive demands that contribute to fatigue and performance degradation over time~\cite{joshi2020}.

Autonomous navigation can reduce this burden, but crowd navigation introduces distinct difficulties. Collisions are most frequent in crowded settings due to the need to anticipate pedestrian trajectories, respect social spaces~\cite{helbing1995, xu2026, kappel2023}, and interpret social context such as ongoing conversations, yielding concerns over safety. Moreover, users' trust in autonomous systems is hindered by lack of transparent communication and hence confusion regarding decision-making~\cite{zhang2021}.

\section{RELATED WORK}

\subsection{Current Autonomous Wheelchair Navigation}

The first intelligent wheelchair was designed with sensors and control for obstacle avoidance by Madarasz in the early 1980's. Since then, controllers have included model-driven methods like PID and MPC, as well as data-driven ones including supervised learning for intent recognition and reinforcement learning. While achieving autonomous navigation, they face limitations in adaptability to dynamic situations and consideration for user autonomy through transparency~\cite{gupta2025}.

Xu et al.\ proposed incorporating user preference fields and a 3D LiDAR informed local planner aware of the wheelchair user and pedestrian's social area, achieving better safety, efficiency, and social rule adherence~\cite{xu2026}. Kappel et al.\ also implemented 3D LiDAR sensing and proxemics-based awareness with corridor detection and lanes, but the system was not tested for dynamic environments yet such as walking pedestrians~\cite{kappel2023}. While addressing social awareness, neither implement user transparent decision-making.

In contrast, Zhang et al.\ improved transparency with shared eHMI, an on-floor projection of the wheelchair's intended direction with short descriptions, where both users and pedestrians reported higher perceived safety, comfort, and trust~\cite{zhang2024ehmi}. Moreover, Guo et al.\ demonstrated through virtual twin simulations that displayed information needs to be carefully curated to minimize distraction, decrease workload, and promote trust while still providing the user enough context~\cite{guo2025}. Nonetheless, these systems only communicate \textit{what} the wheelchair will do, not \textit{why}.

\subsection{Social and Human Behavior Models}

The Social Force Model (SFM) by Helbing and Molnar assigns each pedestrian a circular force-like field based on the shortest path to their destination, desired kinematic states, and need for personal space. These fields can be repulsive or attractive, with magnitude dependent on proximity and directionality. First presented in 1995, this model represented observed pedestrian patterns well, contributing to many adaptations for modeling social situations~\cite{helbing1995}. However, SFM does not reason why pedestrians behave the way they do.

Jha et al.\ proposed \textit{Representing Others’ Trajectories as Executables} (ROTE), which leverages a Large Language Model (LLM) to synthesize behavioral programs' hypothesis space and Sequential Monte Carlo to reason about relative uncertainty. By avoiding behavior cloning or inverse planning, this algorithm avoids large datasets, limited generalizability, and expensive computations, while delivering accuracy similar to human performance~\cite{jha2025}.

\subsection{Problem Definition}

While wheelchairs, particularly powered ones, enable individuals with mobility disabilities to travel locally, their control and the associated informational processes yield a high cognitive load on the user. To overcome this, autonomous or intelligent robotic wheelchairs are being explored to aid users in navigating local environments, but they face barriers to smooth navigation of social environments and trust from the user. Prior work suggests that this could be countered via incorporating both social awareness and transparency in autonomous navigation decisions. 

Our approach draws on the concept proposed by Jha et al. to model other agents’ minds by synthesizing behavioral programs, simulating them forward, and tracking beliefs of each \cite{jha2025}. Instead of directly producing executable code snippets to predict future actions, we produce high level intent hypotheses that will be used to direct future actions. In this work, we present the SAATT Nav framework, Socially Aware Autonomous Transparent Transportation Navigation framework, hereafter referred to as SAATT, that is efficient, safe, and socially aware.
 
\section{Method}

\subsection{System Overview}

\begin{figure*}[b]
\centering
\includegraphics[width=\textwidth]{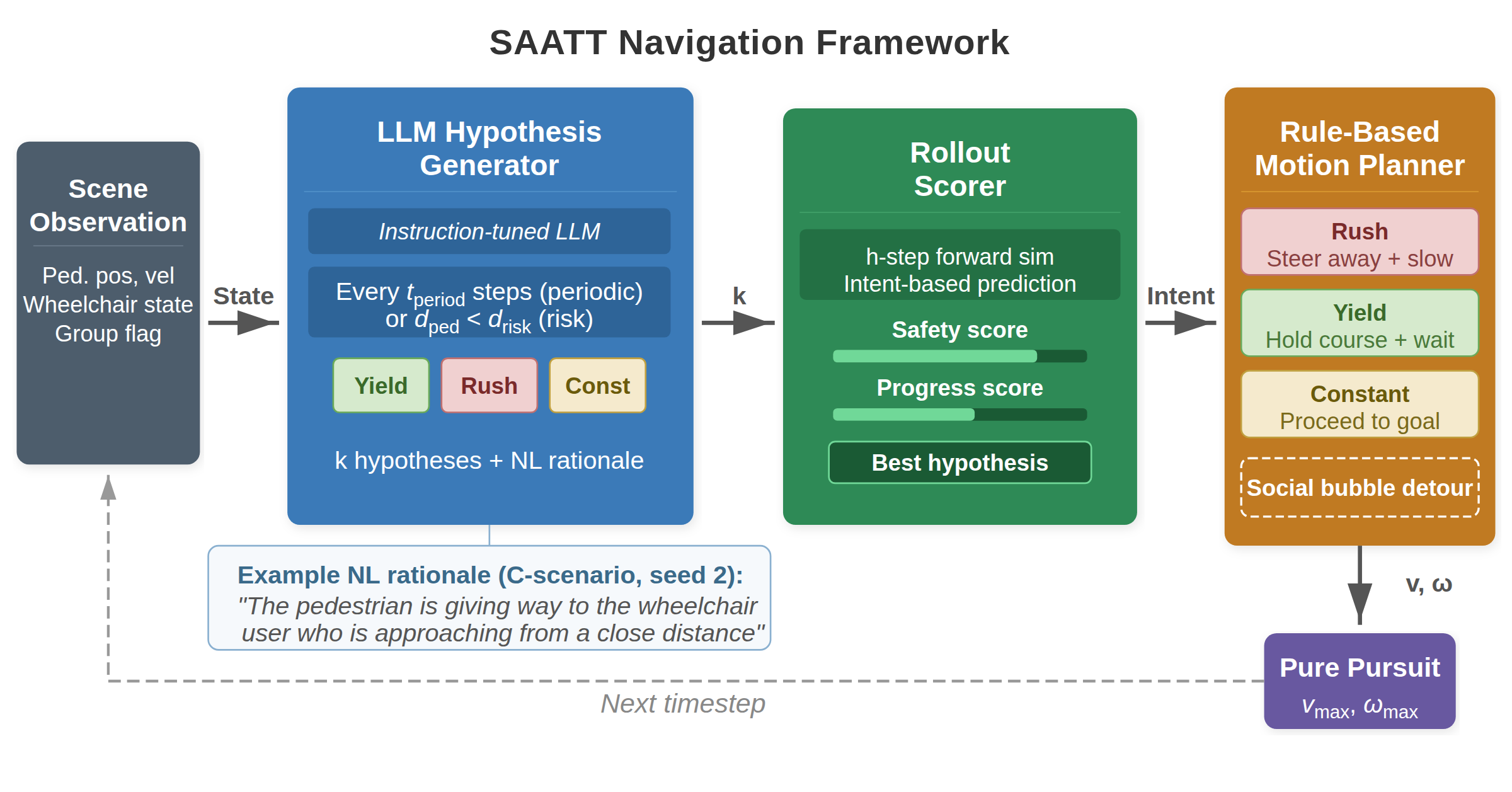}
\caption{Overview of the SAATT Navigation Framework.}
\label{fig:framework}
\end{figure*}

The proposed framework consists of three stages as shown in \cref{fig:framework}. First, an LLM-based hypothesis generator is used to produce a finite set of pedestrian intents. The hypotheses are formed based on the interpretation of the current environment, including wheelchair and pedestrian states, obstacles, and goal position. Each hypothesis categorizes every visible pedestrian based on their predicted behavior and pairs it with a rationale in natural language. A rollout-based scorer then simulates each hypothesis forward over a fixed horizon to evaluate both safety and progress towards goal. The highest-scoring hypothesis is passed to a rule-based motion planner for execution.  

\subsection{LLM-Based Hypothesis Generator }

An LLM model is used to generate pedestrian intent hypotheses each time the trigger condition is met from either of the two scheduling policies. The \textit{periodic} trigger invokes the LLM every $t_{\text{period}}$ timesteps to maintain situational awareness; while the \textit{risk} trigger invokes it when the distance from any pedestrian falls below a threshold $d_{\text{risk}}$ to reduce reaction latency in close encounters. The next invocation time $t_{\text{next}}$ is determined by: 
\begin{equation}\label{eq:trigger}
  t_{\text{next}} = \min\!\left(\,t_{\text{periodic}},\;
  t_{\text{risk}}\,\right),
\end{equation}
where
\begin{flalign}
  &t_{\text{periodic}} = t_{\text{last}} + t_{\text{period}},&
  \label{eq:t_periodic} \\[4pt]
  &t_{\text{risk}} = \min\!\bigl\{\,t > t_{\text{last}}
  + \Delta t_{\min} :&\notag\\
  &\qquad\qquad \min_{i}\lVert \mathbf{p}_{i}(t) - \mathbf{p}_{\mathrm{wc}}(t)
  \rVert < d_{\text{risk}}\,\bigr\}.&
  \label{eq:t_risk}
\end{flalign}
where $t_{\text{last}}$ is the timestep of the most recent LLM call, $\mathbf{p}_{i}(t)$ is the position of pedestrian~$i$, $\mathbf{p}_{\mathrm{wc}}(t)$ is the wheelchair position, and $\Delta t_{\min}$ is the minimum timesteps enforced between consecutive calls to bound inference latency.

Each time the LLM is called, it generates up to $k$ hypotheses to assign one of the three behavioral labels (\textit{yield}, \textit{rush}, or\textit{ constant velocity}) to each pedestrian in the environment, along with a rationale behind the chosen intent in natural language. The sampling temperature $\tau$ is chosen to balance creativity and coherence in the generated output.

\subsection{Rollout-Based Hypothesis Scorer}

After each set of $k$ hypotheses is generated, each one is evaluated through a forward rollout simulation over a horizon of $h$ timesteps. During each rollout, the wheelchair moves toward the goal under a pure pursuit controller, while the pedestrians' velocities are modulated according to the predicted intent. Let $\mathbf{v}_{i}$ denote the observed velocity of pedestrian~$i$; the
simulated velocity $\hat{\mathbf{v}}_{i}$ is:
\begin{equation}\label{eq:ped_velocity}
  \hat{\mathbf{v}}_{i} =
  \begin{cases}
    \alpha_{\text{yield}}\;\mathbf{v}_{i}
      & \text{if intent} = \textit{yield}, \\
    \alpha_{\text{rush}}\;\mathbf{v}_{i}
      & \text{if intent} = \textit{rush}, \\
    \mathbf{v}_{i}
      & \text{if intent} = \textit{constant velocity},
  \end{cases}
\end{equation}
where $\alpha_{\text{yield}} < 1$ and $\alpha_{\text{rush}} > 1$ are scaling factors chosen so that the three intents produce meaningfully distinct rollout trajectories. Since hypotheses are ranked on a relative scale, the exact magnitude of the scaling factors does not affect the final selection.

The rollout results are used to compute both a safety score and a progress score for each hypothesis. Let $n_{\text{coll}}$ denote the number of timesteps in which a collision occurs during the rollout, $d_{\min}$ the minimum wheelchair-to-pedestrian distance observed over the horizon, and $d_0$, $d_f$ the wheelchair-to-goal distance at the start and end of the rollout respectively. The scores are defined as:
\begin{equation}\label{eq:safety}
  S_{\text{safety}} =
  \begin{cases}
    \max\!\bigl(0,\;1 - \frac{n_{\text{coll}}}{h}\cdot 2\bigr)
      & \text{if } n_{\text{coll}} > 0, \\[4pt]
    \min\!\bigl(1,\;\frac{d_{\min}}{d_{\text{safe}}}\bigr)
      & \text{if } n_{\text{coll}} = 0 \\
      & \text{and } d_{\min} < d_{\text{safe}}, \\[4pt]
    1
      & \text{otherwise},
  \end{cases}
\end{equation}
\begin{equation}\label{eq:progress}
  S_{\text{progress}} = \max\!\left(0,\;\frac{d_0 - d_f}{d_0}\right).
\end{equation}
A collision is registered at any timestep where
$\lVert \mathbf{p}_{i} - \mathbf{p}_{\mathrm{wc}} \rVert < r_{\mathrm{wc}} + r_{\mathrm{ped}}$,
the sum of the wheelchair and pedestrian radii.
The two scores are combined into a total score:
\begin{equation}\label{eq:total}
  \begin{split}
    S_{\text{total}} = \; & w_p \, S_{\text{progress}} + w_s \, S_{\text{safety}} \\
    & - \lambda_c \, n_{\text{coll}} - \lambda_b \,\mathbb{1}[S_{\text{social}} < 0]
  \end{split}
\end{equation}
where $w_s > w_p$ so that safety takes priority over progress, $\lambda_c$ and $\lambda_b$ are penalty weights for collision and social bubble intrusion respectively, and $\mathbb{1}[\cdot]$ is the indicator function. The hypothesis with the highest $S_{\text{total}}$ is selected and its predicted intent is forwarded to the motion planner for execution.

\subsection{Rule-Based Motion Planner}

The motion planner translates the selected hypothesis into a control command for the wheelchair. The predicted pedestrian intent determines which reactive strategy is applied, and the speed limit within each strategy depends on the current distance to the nearest pedestrian
$d_{\text{ped}}$.

Under \textit{constant velocity}, the wheelchair proceeds toward the goal at its nominal speed $v_{\text{nom}}$. Under \textit{rush} and \textit{yield}, the wheelchair progressively reduces speed based on the current distance to the nearest pedestrian $d_{\text{ped}}$:
\begin{equation}\label{eq:v_ref}
  v_{\text{ref}} =
  \begin{cases}
    0             & \text{if } d_{\text{ped}} < d_{\text{stop}}^*, \\
    v_{\text{slow}}^* & \text{if } d_{\text{ped}} < d_{\text{slow}}^*, \\
    v_{\text{nom}} & \text{otherwise},
  \end{cases}
\end{equation}
where $d_{\text{stop}}^*$, $d_{\text{slow}}^*$, and
$v_{\text{slow}}^*$ are intent-dependent thresholds. The
\textit{yield} thresholds are set larger than those for \textit{rush}
such that the wheelchair begins decelerating earlier when it expects the pedestrian to slow down. Under \textit{rush}, the wheelchair additionally steers laterally away from the pedestrian.  

When the LLM module is active and a social group is detected, defined as two or more stationary pedestrians in close proximity, the planner computes a detour waypoint around the group's social bubble. The bubble is centered at the midpoint of the two pedestrians with radius:
\begin{equation}\label{eq:bubble}
  r_{\text{bubble}} = \tfrac{1}{2}\,\lVert \mathbf{p}_{1} - \mathbf{p}_{2}
  \rVert + r_{\text{ped}} + m_{\text{personal}} + m_{\text{group}},
\end{equation}
where $\mathbf{p}_{1}$, $\mathbf{p}_{2}$ are the pedestrian
positions, $r_{\text{ped}}$ is the pedestrian radius,
$m_{\text{personal}}$ is the personal space margin, and
$m_{\text{group}}$ is the group space margin. Even when the physical gap between the two pedestrians is wide enough for the wheelchair to pass through, the planner treats the conversational space as a unified social zone and routes around it. The detour relies on the LLM's ability to recognize the pedestrian group as a social interaction rather than two independent obstacles; without hypothesis generation, the planner has no basis to distinguish a conversational pair from unrelated bystanders and therefore does not trigger the detour.

At the low level, a pure pursuit controller converts each planner waypoint into linear and angular velocity commands:
\begin{align}
  \omega &= k_{\theta}\;\text{wrap}\!\bigl(\theta_{\text{desired}}
  - \theta\bigr), \label{eq:omega}\\[4pt]
  a &= k_{v}\;\bigl(v_{\text{ref}} - v\bigr), \label{eq:accel}
\end{align}
where $\theta_{\text{desired}} = \text{atan2}(y_{\text{wp}} - y,\;
x_{\text{wp}} - x)$ is the heading toward the waypoint,
$\text{wrap}(\cdot)$ normalizes the angle to $[-\pi,\pi]$,
and $k_{\theta}$, $k_{v}$ are proportional gains. Both $\omega$ and
$a$ are clipped to $[\omega_{\min},\omega_{\max}]$ and
$[a_{\min},a_{\max}]$ respectively.

The complete procedure is summarized in Algorithm~\ref{alg:saatt}.

\begin{algorithm}
\small  
\caption{SAATT Navigation Framework}
\label{alg:saatt}
\begin{algorithmic}[1]
\Require config, goal position $\mathbf{g}$, max steps $N$
\State Initialize environment, planner, controller
\State $t_{\text{last}} \leftarrow -\infty$, $\mathcal{H} \leftarrow \emptyset$
\For{$t = 0$ \textbf{to} $N$}
    \State Observe $\mathbf{s}_{\text{wc}} = (x, y, \theta, v)$, pedestrians $\{\mathbf{p}_i, \mathbf{v}_i\}$
    \If{$t \geq t_{\text{last}} + t_{\text{period}}$ \textbf{or} $\min_i \| \mathbf{p}_i - \mathbf{p}_{\text{wc}} \| < d_{\text{risk}}$}
        \If{$t - t_{\text{last}} \geq \Delta t_{\min}$}
            \State $\mathcal{H} \leftarrow \text{LLM.generate}(k \text{ hypotheses})$
            \State $t_{\text{last}} \leftarrow t$
        \EndIf
    \EndIf
    \For{\textbf{each} $h_j \in \mathcal{H}$}
        \State Simulate $h$ steps with $\hat{\mathbf{v}}_i$ per intent
        \State Compute $S_{\text{safety}}^j$, $S_{\text{progress}}^j$, $S_{\text{total}}^j$
    \EndFor
    \State $h^* \leftarrow \arg\max_j\; S_{\text{total}}^j$
    \State $(\mathbf{w}, v_{\text{ref}}) \leftarrow \text{Planner}(\text{obs},\; h^*)$
    \State $\omega \leftarrow k_\theta \;\text{wrap}(\theta_{\text{desired}} - \theta)$
    \State $a \leftarrow k_v \;(v_{\text{ref}} - v)$
    \State Apply $(a, \omega)$ to environment
    \If{goal reached \textbf{or} collision}
        \State \textbf{break}
    \EndIf
\EndFor
\end{algorithmic}
\end{algorithm}

\subsection{Baseline Methods}

As a global planner, the A* baseline computes the shortest path on a discretized grid and follows it using the same pure pursuit controller. Pedestrians are treated as inanimate obstacles with no proxemics or social bubbles in the occupancy grid, so the planner produces the same path regardless of the social context.

The SFM baseline is inspired by the social force framework of Helbing and Moln\'{a}r~\cite{helbing1995}, adapted for wheelchair navigation. The net force acting on the wheelchair is:
\begin{equation}\label{eq:sfm}
  \mathbf{F} = \mathbf{F}_{\text{goal}}
  + \sum_{i} \mathbf{F}_{\text{ped},i}
  + \sum_{j} \mathbf{F}_{\text{wall},j},
\end{equation}
where $\mathbf{F}_{\text{goal}}$ is an attractive force directed toward the goal, and $\mathbf{F}_{\text{ped},i}$,
$\mathbf{F}_{\text{wall},j}$ are repulsive forces that decay exponentially with distance from pedestrian~$i$ and wall~$j$ respectively. The net force is normalized to produce a short-horizon subgoal tracked by the pure pursuit controller. Because SFM treats each pedestrian independently, it has no notion of a shared conversational space and cannot avoid social bubbles.

The Ablation method uses the same rule-based planner as the proposed SAATT framework but with the LLM module disabled. Without LLM inference, no hypothesis is generated, no intent-based speed modulation is applied, and no social bubble detour is triggered. Any performance
difference between the proposed framework and the Ablation is therefore attributable to LLM-based reasoning alone.

\section{EXPERIMENTAL SETUP}

\subsection{Simulation Environment}
All experiments were conducted in a custom 2D simulation environment implemented in Python, modeling a 10m-by-10m space. The wheelchair is represented as a circular agent with a radius of 0.4 m, a maximum linear velocity of 1.5 m/s \cite{cooper2002}, and differential drive kinematics \cite{gersdorf2010}. Pedestrians are modeled as circular agents with a radius of 0.3 m. A collision is registered when the distance between the wheelchair center and any pedestrian center falls below 0.7 m, corresponding to the sum of their radii. Each trial runs at a timestep of 0.1 s for a maximum of 1000 steps (100 s) and terminates early if the wheelchair reaches within 0.5 m of the goal. 

The number of hypotheses $k=8$ and minimum interval $\Delta t_{\min}=15$ steps between LLM calls were chosen to limit inference latency for eventual deployment on embedded hardware. The sampling temperature $\tau=0.7$ balances creative reasoning and coherence, as values above 1.0 significantly degrade causal reasoning and instruction following~\cite{renze2024, li2025}. The rollout speed scales $\alpha_{\text{yield}}=0.2$ and $\alpha_{\text{rush}}=1.5$ produce distinct trajectories for scoring; since hypotheses are ranked on a relative scale, the exact magnitudes do not affect the final selection. The safe distance threshold $d_{\text{safe}}=2.0$\,m lies within the social distance zone of 1.2--3.6\,m identified in~\cite{rios2015}. Safety is weighted twice as heavily as progress ($w_s=20, w_p=10$) to prioritize collision avoidance over travel efficiency, consistent with the safety-first principle in human-aware navigation~\cite{kruse2013}.

All system parameters are summarized in \cref{tab:params}.

\subsection{Scenarios}

Three scenarios of increasing social complexity were designed to evaluate the proposed framework. 

Scenario A features a open corridor with no pedestrians. 
The wheelchair navigates from a randomized start position to a randomized goal. This scenario aims to verify that all four methods produce equivalent performance in the absence of social context. 

Scenario B consists of a conversing pair where two stationary pedestrians are placed in a face-to-face arrangement, forming an F-formation \cite{kendon1990}, or social bubble. The wheelchair navigates from start to goal in the presence of this social bubble. 

Scenario C includes a crossing pedestrian that walks horizontally across the wheelchair path at a randomized speed between 0.4 m/s and 0.8 m/s. This scenario tests the ability to predict pedestrian intent and respond appropriately to a dynamic agent.

\begin{table}[hbt!]
\centering
\caption{System Parameters}
\label{tab:params}
\begin{tabular}{@{}llr@{}}
\toprule
\textbf{Symbol} & \textbf{Description} & \textbf{Value} \\
\midrule
\multicolumn{3}{@{}l}{\textit{Environment}} \\
$r_{\mathrm{wc}}$ & Wheelchair radius & 0.4\,m \\
$r_{\mathrm{ped}}$ & Pedestrian radius & 0.3\,m \\
$v_{\max}$ & Max linear velocity & 1.5\,m/s \\
$\Delta t$ & Simulation timestep & 0.1\,s \\
\midrule
\multicolumn{3}{@{}l}{\textit{LLM Hypothesis Generator}} \\
$t_{\text{period}}$ & Periodic trigger interval & 30 steps \\
$d_{\text{risk}}$ & Risk trigger distance & 3.0\,m \\
$\Delta t_{\min}$ & Min steps between calls & 15 steps \\
$k$ & Hypotheses per call & 8 \\
$\tau$ & Sampling temperature & 0.7 \\
\midrule
\multicolumn{3}{@{}l}{\textit{Rollout Scorer}} \\
$h$ & Rollout horizon & 30 steps \\
$\alpha_{\text{yield}}$ & Yield speed scale & 0.2 \\
$\alpha_{\text{rush}}$ & Rush speed scale & 1.5 \\
$d_{\text{safe}}$ & Safe distance threshold & 2.0\,m \\
$w_p$ & Progress weight & 10.0 \\
$w_s$ & Safety weight & 20.0 \\
$\lambda_c$ & Collision penalty & 50.0/step \\
$\lambda_b$ & Bubble penalty & 100.0 \\
\midrule
\multicolumn{3}{@{}l}{\textit{Motion Planner}} \\
$v_{\text{nom}}$ & Nominal speed & 1.5\,m/s \\
$d_{\text{slow}}$ & Slow-down radius & 1.0\,m \\
$d_{\text{stop}}$ & Stop radius & 0.6\,m \\
$m_{\text{bubble}}$ & Social bubble margin & 0.3\,m \\
\midrule
\multicolumn{3}{@{}l}{\textit{Pure Pursuit Controller}} \\
$k_{\theta}$ & Heading gain & 2.0 \\
$k_{v}$ & Speed gain & 1.0 \\
$\omega_{\max}$ & Max angular velocity & 1.0\,rad/s \\
\bottomrule
\end{tabular}
\end{table}

\subsection{Randomization and Paired Design}
For each scenario, 30 unique layouts were generated using random seeds. Each seed determines the start position, goal position, and pedestrian placement. All four methods were evaluated on the exact same 30 layouts per scenario, creating a paired experimental design that ensures performance differences are not confounded by layout variation. The total number of trials is 3 scenarios by 4 methods by 30 seeds, yielding 360 experiments.

\subsection{Performance Metrics}
Following evaluation frameworks for socially-aware robot navigation \cite{karwowski2023, gao2022, kruse2013}, we assess performance across four categories: Safety, Social Compliance, Efficiency, and Comfort. Safety is measured by the Collision Count (total timesteps in violation) and the Minimum Pedestrian Distance (closest approach to any pedestrian across an entire trial). Social Compliance in Scenario B is measured by Bubble Intrusion Steps, the number of timesteps the wheelchair spent inside the social bubble. Efficiency is measured by Travel Time, Path Length, and Success in goal-reaching. Comfort is assessed through Mean Angular Velocity and Mean Jerk, which captures smoothness of heading changes and acceleration profiles respectively.

\section{RESULTS AND DISCUSSION}

\subsection{Statistical Analysis}
\begin{figure}[t]
    \centering
    \includegraphics[width=0.9\columnwidth]{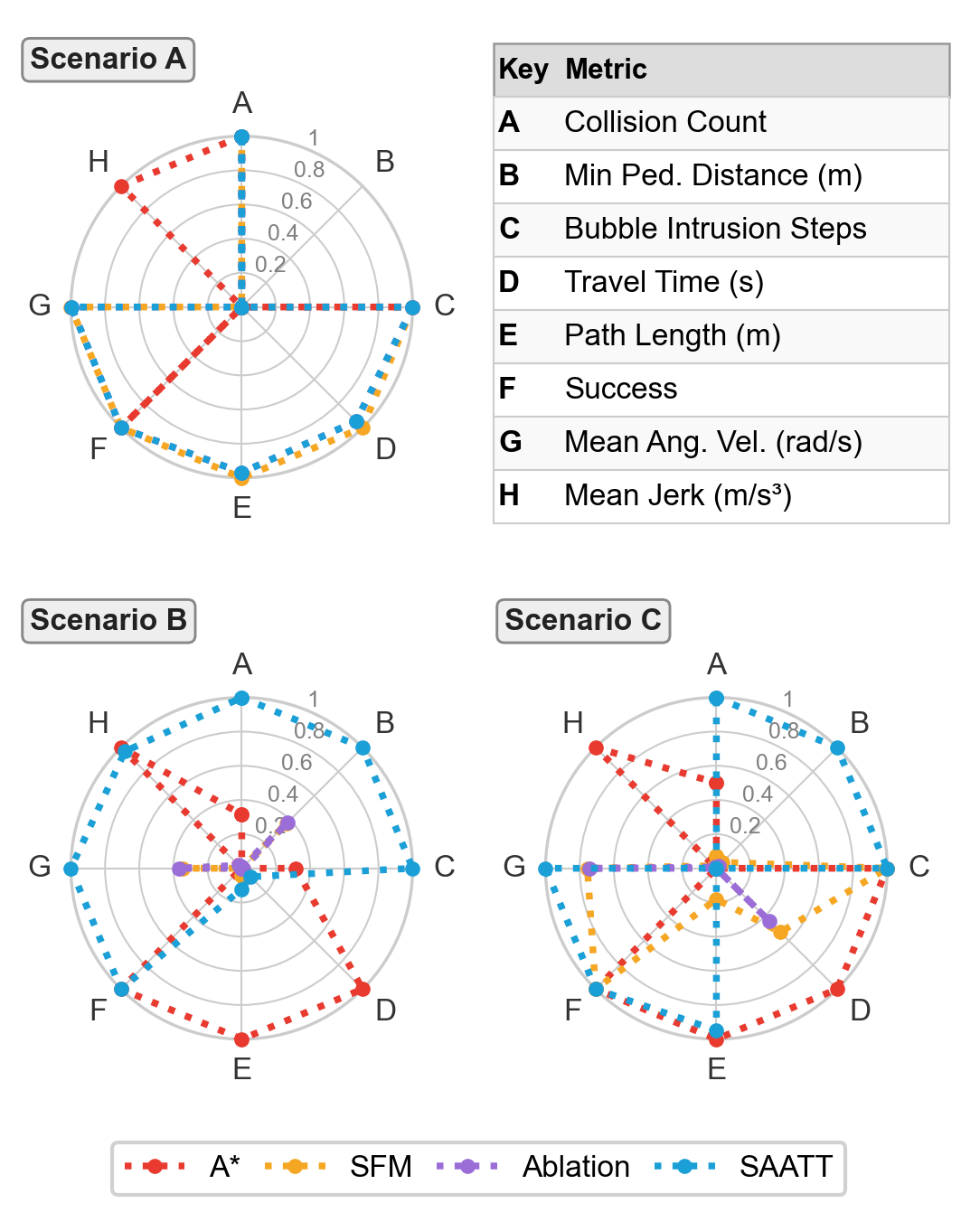}
    \caption{Radar plots for performance metrics across scenarios, normalized via Min-Max Scaling. Complements were found for all but Success and Minimum Pedestrian Distance to ensure the outer ring correlates to better performance.}
    \label{fig:radar_vertical}
\end{figure}

\begin{figure*}[t]
\centering
\includegraphics[width=\textwidth]{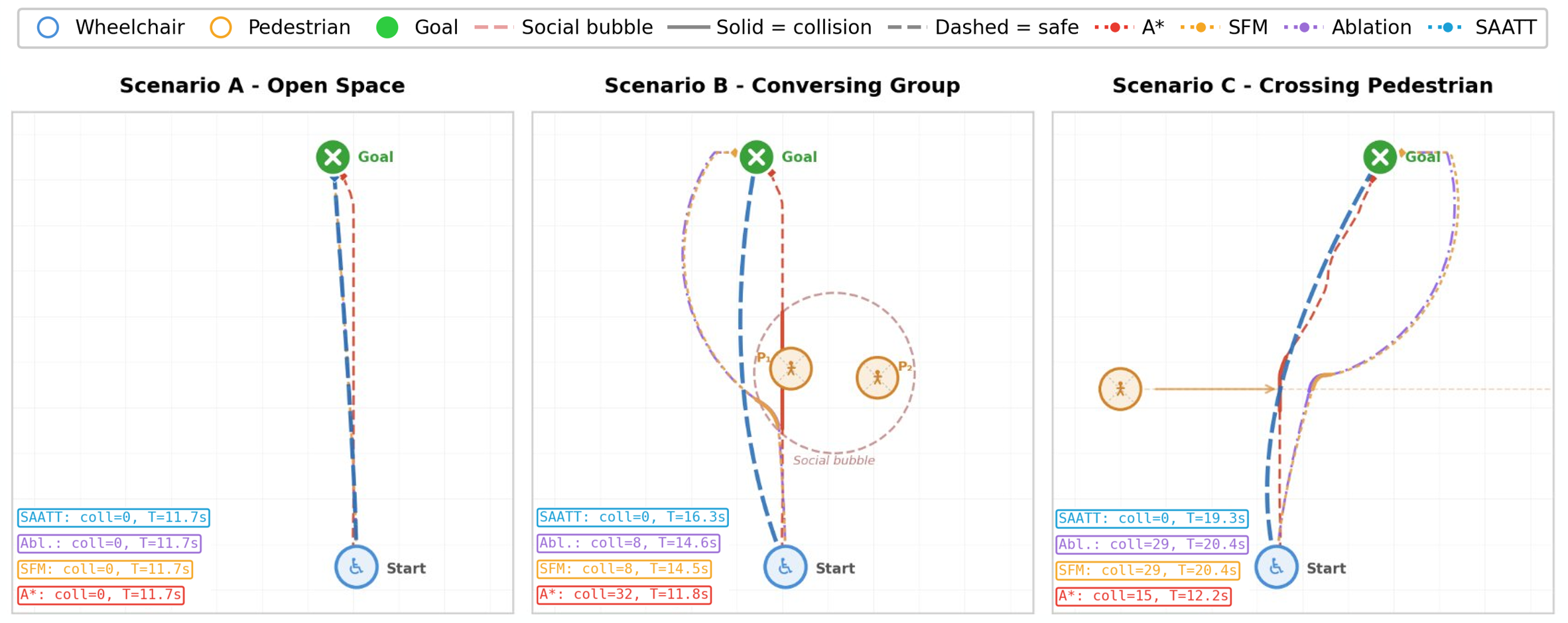}
\caption{Trajectory comparison across all four methods in Scenario A (left), B (center), and C (right) for a representative seed.}
\label{fig:overlay}
\end{figure*}

Every metric besides successes, was non-binary data, containing 120 total experimental trials including 30 randomized seeds matched across all navigation methods for each environment. A normalized visualization summarizing every metric's average is in \cref{fig:radar_vertical}. An example comparison of the same seed across all three scenarios, including Collision Count, Travel Time, and Path Length is shown in \cref{fig:overlay}. Since they had the same experimental treatment, the data was considered dependent. However, it was unknown if the data was parametric or not, so the Shapiro-Wilk Normality test was used to ensure the correct statistical analysis would be performed for each method and environment combination (12 total). Ultimately, majority of the metrics’ p-values were less than 0.05, meaning they failed the tests and hence were considered abnormal distributions. 

As the data was deemed nonparametric, paired, and dependent, the Friedman’s test was conducted to determine if any of the methods yielded statistically significant difference compared to the other groups for each scenario when p-value is less than 0.05. If one was detected, the Conover test was then utilized to perform paired t-test comparisons between each of the methods and rank the best performers due to its high power and ability to detect real differences. These statistical significances or lack thereof and ranking results are shown in \cref{tab:result}.

The final metric is Success, which consists of digital values for each trial that was performed. Due to the data’s binary nature, it is non-parametric. In order to detect statistically significant difference in success rates Cochran’s Q Test was used for each scenario environment. Due to the near 100 percent success rates for each method per scenario, the p-values were incalculable or were greater than 0.05, communicating that there was no statistically significant differences in path completion. 

\subsection{Safety}
 Scenario A had no statistically significant differences for both Collision Count and Minimum Pedestrian Distance. In other terms, all navigation methods took relatively similar or same path in the first case due to the lack of human or dynamic obstacles. 
 
 SAATT had the least number of collisions in both scenario B and C. For Minimum Pedestrian Distance, the largest minimum distance was achieved by SAATT in scenario B and C again. This suggests that the SAATT model is best at avoiding collisions and maintaining space between the wheelchair user and surrounding pedestrians when navigating social and dynamic environments, which correlates with better safety for the user and pedestrians.

\subsection{Social Compliance}
The number of timesteps of the wheelchair user in the social bubble, Bubble Intrusion Steps, was only observed in scenario B. In the SAATT Nav framework, the wheelchair spent the least timesteps in the social bubble, avoiding encroaching on the social interaction compared to the other three methods.  

\subsection{Efficiency}
For Travel Time, A* completed their trajectories  in the shortest time periods for Scenarios B and C, while SAATT performed better for Scenario A. This better performance was likely due to the fact that A* ignored social situations, while SAATT took it into consideration. It is worth noting that the range for this dataset is 2.3 seconds while it is 88.4 and 13.4 seconds for scenario B and C respectively, meaning that scenario A’s times are relatively similar. Also, SAATT had statistically similar performances with Ablation and SFM methods.

The performance across the scenarios for Path Lengths was much more varied with the best being SFM for Scenario A, A* for Scenario B, and SAATT for Scenario C. Also, SAATT performed similarly to Ablation in Scenario A. The first scenario had a data range of 1.885\,m, while scenario B and C had 10.87\,m and 4.043\,m respectively. For B, A* likely had the shortest lengths as it ignored the social bubble and navigated between both human obstacles. A and C had relatively low Path Lengths ranges so the differences may not be as important but SAATT performing better with the introduction of the dynamic social element is worth highlighting. 

As mentioned in the statistical analysis section, the success rates yielded no statistically significant differences. While this does not support any claims, it does demonstrate that all methods are able to properly navigate to the goal. 

\begin{table*}[ht]
\centering
\arrayrulecolor{black}
\caption{Performance comparison across scenarios A, B, and C.}
\label{tab:result}
\resizebox{\textwidth}{!}{%
\begin{tabular}{|l|l|l|l|l|}
\hline
\textbf{Category} & \textbf{Metric} & \textbf{A} & \textbf{B} & \textbf{C} \\
\hline

\multirow{2}{*}{\textbf{Safety}}
  & Collision Count (low to high)
  & \cellcolor{lightgray}
  & \cellcolor{lightgreen} SAATT, Ablation/SFM, A*
  & \cellcolor{lightgreen} SAATT, A*, SFM/Ablation \\
\hhline{~----}
  & Minimum Pedestrian Distance (m) (high to low)
  & \cellcolor{lightgray}
  & \cellcolor{lightgreen} SAATT, Ablation/SFM, A*
  & \cellcolor{lightgreen} SAATT, SFM, A*/Ablation \\
\hline

\textbf{Social Compliance}
  & Bubble Intrusion Steps (low to high)
  & \cellcolor{lightgray}
  & \cellcolor{lightgreen} SAATT, Ablation/SFM, A*
  & \cellcolor{lightgray} \\
\hline

\multirow{3}{*}{\textbf{Efficiency}}
  & Travel Time (s) (low to high)
  & \cellcolor{lightgreen} \textbf{Ablation/SAATT}/SFM, A*
  & \cellcolor{lightred} A*, SFM/Ablation, SAATT
  & \cellcolor{lightred} A*, SFM/Ablation, SAATT \\
\hhline{~----}
  & Path Length (m) (low to high)
  & \cellcolor{lightred} SFM, \textbf{SAATT/Ablation}, A*
  & \cellcolor{lightred} A*, SAATT, Ablation/SFM
  & \cellcolor{lightgreen} SAATT, SFM/A*/Ablation \\
\hhline{~----}
  & Success Rate (N/A)
  & \cellcolor{lightgray}
  & \cellcolor{lightgray}
  & \cellcolor{lightgray} \\
\hline

\multirow{2}{*}{\textbf{Comfort}}
  & Mean Angular Velocity (rad/s) (low to high)
  & \cellcolor{lightred} SFM, \textbf{Ablation/SAATT}, A*
  & \cellcolor{lightgreen} SAATT, Ablation/A*/SFM
  & \cellcolor{lightgreen} \textbf{SAATT/Ablation}, SFM, A* \\
\hhline{~----}
  & Mean Jerk (m/s$^3$) (low to high)
  & \cellcolor{lightred} A*, SFM/\textbf{SAATT/Ablation}
  & \cellcolor{lightgreen} A*/SAATT, Ablation, SFM
  & \cellcolor{lightred} A*, \textbf{Ablation/SAATT}/SFM \\
\hline

\end{tabular}%
}
\vskip 6pt
\parbox{\linewidth}{\fontsize{8}{10}\selectfont \textit{Note:} The gray boxes represent no statistically significant difference across all groups, green boxes indicate where SAATT is ranked first, and red boxes indicate when they are not. Methods separated by ``/'' are deemed not statistically significantly different. Combinations of SAATT and Ablation are bolded as well.}
\end{table*}

\subsection{Comfort}
The lowest Mean Angular Velocity metrics were achieved by SFM for Scenario A, and SAATT for Scenarios B and C. Additionally, the data demonstrated similarity between SAATT and the Ablation for Scenarios A and C. The higher ranks for the two social situations show that SAATT was better at avoiding rapid, sharp turns, contributing to a smoother trajectory. 

A* achieved the lowest Mean Jerk for all scenarios with SAATT having no difference for Scenario B. This is most likely due to the global planner being designed to take the shortest path, making it inherently ignore the social situations. Again, it is also worth noting that the data ranges for scenarios A, B, and C are 0.0684, 0.2376, and 0.3906 seconds respectively, all of which are really small and may be negligible in real world applications. This means that all methods have no major fluctuations in acceleration that cause the rider to be uncomfortably lurched by the wheelchair movement.

\subsection{LLM vs. Ablation Performance}
Although there are some cases where SAATT and Ablation had the same behavior, such as all metrics for scenario A and specifically Efficiency, Comfort, and Social Compliance categories for scenario C, this behavior was not universal. First, in these cases the performance was most likely deemed statistically similar, not exactly the same, due to using the same local planner  responsible for base functionality. Second, the SAATT method differed from the Ablations in all metrics for scenario B and specifically Safety metric category for scenario C, which is confirming as Safety and Scenario B were more tied to the pedestrians themselves. Since the only difference between the two methods is the LLM incorporating pedestrian intent, these varied demonstrated behaviors supports that intent-prediction does influence its decision making.

\begin{figure}[!b]
\centering
\includegraphics[width=\columnwidth]{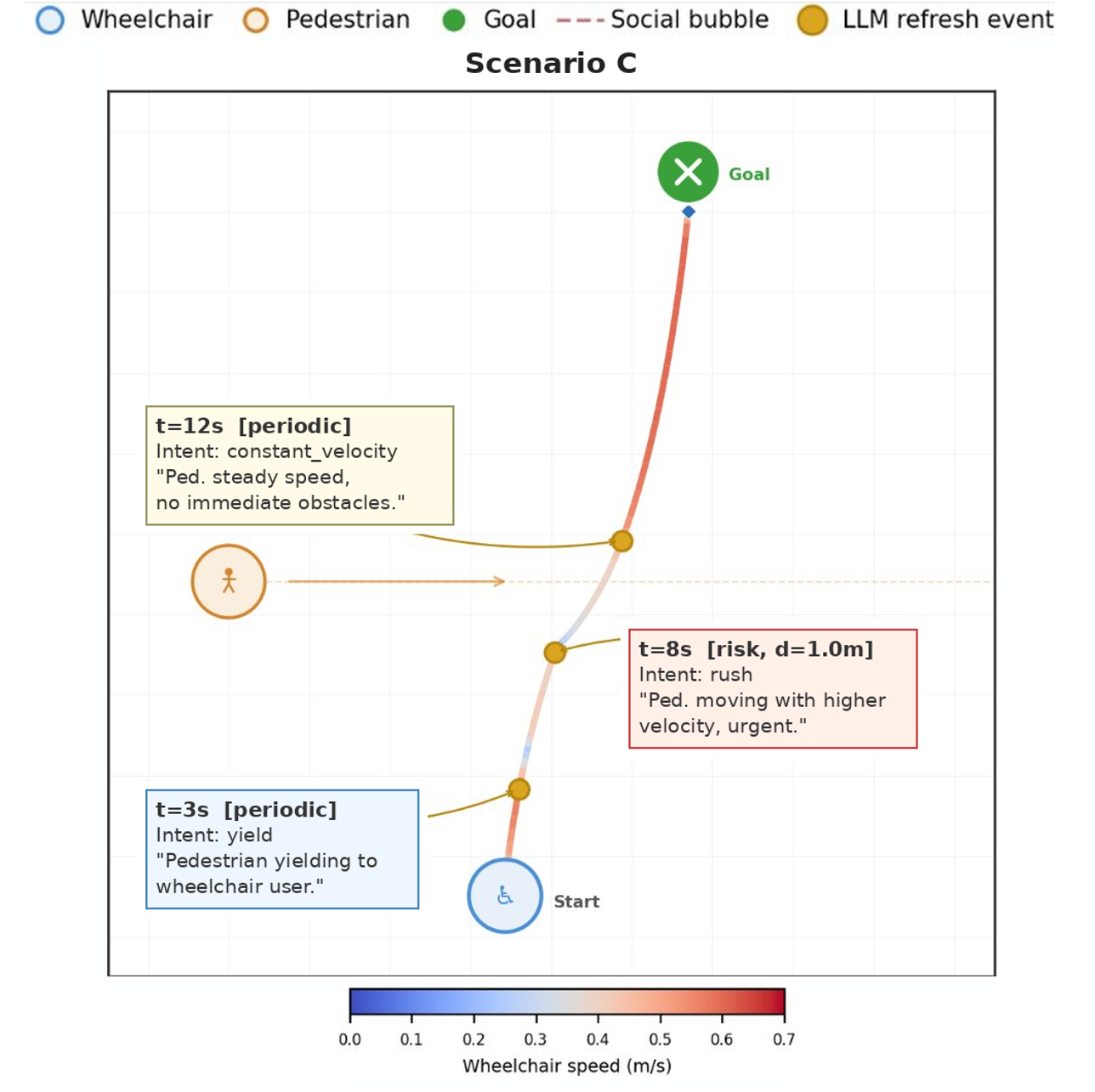}
\caption{Qualitative results of the proposed framework in Scenario C, showing wheelchair trajectory, LLM refresh events, and predicted pedestrian intents to demonstrating transparency.}
\label{fig:rationale}
\end{figure}

\subsection{Transparency}
As demonstrated in \cref{fig:rationale}, the LLM system is capable of generating messages along the autonomous wheelchair trajectory. The message communicates the navigation systems' intentions and logic or reducing user confusion and distrust. Research in social sciences shows that people prefer selective, cause-focused explanations over exhaustive causal chains. This transforms the otherwise black-box navigation into a transparent decision-making process \cite{miller2019}.

\section{LIMITATIONS AND FUTURE DIRECTIONS}
Currently, SAATT treats the humans in social situations as repellent forces. While this is applicable to non-attractive strangers, it is possible that the user might want to approach them to interact, which could be represented via attractive forces \cite{helbing1995}. Furthermore, we could provide more context to the LLM that resembles natural social cues in everyday interactions, such as a person’s body language, eye gaze, or vocal cues \cite{adams2017}. In this scenario, if two people are oriented away from each other, the system may infer that they are not engaged in conversation and that it is socially permissible to pass between them.

Another future direction is to incorporate a user preference field where top hypotheses are presented to the user for decision making. This mitigates over-reliance and complacency seen in autonomous driving research \cite{wen2019} while supporting the higher sense of agency preferred by assistive robotics users \cite{collier2025}. The system could be personalized per user via machine learning \cite{ahmadi2026} to build trust. Alternatively, tuning LLM temperature and model size could better align SAATT's social navigation behavior with user preferences, as temperature affects reasoning, creativity, and instruction following~\cite{renze2024, li2025}.

Lastly, as edge-device deployment is explored in real-world settings, inference techniques need to be optimized to meet the latency constraints of real-time navigation. To further real-world translational merit, simulation scenarios evaluated in this work should be extended to higher-density and more complex social dynamics to validate SAATT's robustness across diverse conditions encountered in realistic assistive navigation.

\section{CONCLUSIONS}

By using the LLM as a human-proxy for decision making, the simulations demonstrated SAATT's ability to avoid dynamic and social obstacles, such as strangers conversing  or crossing pedestrians, that a wheelchair user would need to overcome in a more socially acceptable way.  

When compared to the global planner A*, SAATT demonstrated better performance in the metric categories of Safety and Social Compliance, equivalent performance in terms of Success, Path Length for the dynamic social situation, and partially in Comfort specifically Mean Angular Velocity for both social situations. A* did perform better in the remaining metrics in the Efficiency category and Comfort, but it is biased by categorizing individuals as inanimate objects and encroaching personal and social bubbles.  

For SFM, our representative for other social navigation methods, SAATT performed better or equivalent to it for all categories in scenarios B and C. In scenario A, SAATT performed slightly worse in Path Length and Mean Angular Velocity, while equivalently for Travel Time and Mean Jerk. The comparison between Ablation and SAATT also demonstrated the benefits of understanding intent of the social obstacles when navigating social scenarios. 

Finally, the system displayed its reasoning for each waypoint to the user to improve transparency. While the study lacked human users to assess the explanations, its ability to communicate its intentions and logic serves as an example of it potential to develop trust with its user and sense of safety within its user.  

Despite the discussed limitations, this work proposes SAATT, a potential method for autonomous navigation that successfully accounts for social awareness and transparency by incorporating an LLM model for making more human-like social decisions and communicating with its user. 

\addtolength{\textheight}{-15cm}   




\section*{ACKNOWLEDGMENT}

This research was funded, in part, by the Advanced Research Projects Agency for Health (ARPA-H) Agreement No. 140D042590012. The views and conclusions contained in this document are those of the authors and should not be interpreted as representing the official policies, either expressed or implied, of the U.S. Government. This work was partially supported by the Center for AI and Robotic Excellence in medicine (CARE) at Purdue University. Computation in this work used Anvil at Purdue University through allocation cis260101-gpu from the Advanced Cyberinfrastructure Coordination Ecosystem: Services \& Support (ACCESS) program, which is supported by U.S. National Science Foundation grants \#2138259, \#2138286, \#2138307, \#2137603, and \#2138296. The authors also acknowledge the use of Claude (Anthropic) for assistance in manuscript language polishing, code debugging, and the development of data visualization scripts. Additionally, data analysis for this paper was conducted using the Real Statistics Resource Pack software (Release 9.4.5), Copyright (2013 -- 2026) Charles Zaiontz (www.real-statistics.com).

Omitted for anonymous review.

\balance

\bibliographystyle{IEEEtran}
\bibliography{refs}

\end{document}